\pdfoutput=1
\documentclass[11pt]{article}

\usepackage[]{ACL2023}

\usepackage{times}
\usepackage{latexsym}
\usepackage{siunitx}

\usepackage[T1]{fontenc}
\usepackage[utf8]{inputenc}

\usepackage{microtype}

\usepackage{inconsolata}
\usepackage{caption}

\usepackage{graphicx}
\usepackage{multirow}
\usepackage{adjustbox}
\usepackage{tabularx}

\usepackage{enumitem}
\usepackage{amsfonts}

\usepackage{siunitx}
\usepackage{dsfont}
\usepackage{amsmath}
\usepackage{amssymb}
\usepackage{bbm}
\usepackage{pifont}

\usepackage{tabularx}
\usepackage{tablefootnote}
\usepackage[utf8]{inputenc}
\usepackage{subfigure}
\usepackage{parskip}

\usepackage{color}
\usepackage{comment}
\usepackage{booktabs}

\usepackage{makecell}
\usepackage{bbding}
\usepackage{algorithm}
\usepackage{algpseudocode}

\usepackage{courier}
\usepackage{url}

\title{Prompt to be Consistent is Better than Self-Consistent? Few-Shot and Zero-Shot Fact Verification with Pre-trained Language Models}

\author{Fengzhu Zeng \and Wei Gao \\
   School of Computing and Information Systems \\	
    Singapore Management University \\ 
   	80 Stamford Rd, Singapore 178902 \\
   \texttt{fzzeng.2020@phdcs.smu.edu.sg}, 
   \texttt{weigao@smu.edu.sg}}   

\begin{document}
\maketitle
\begin{abstract}
Few-shot or zero-shot fact verification only relies on a few or no labeled training examples. In this paper, we propose a novel method called ProToCo, to \underline{Pro}mpt pre-trained language models (PLMs) \underline{To} be \underline{Co}nsistent, for improving the factuality assessment capability of PLMs in the few-shot and zero-shot settings.
Given a claim-evidence pair, ProToCo generates multiple variants of the claim with different relations and frames a simple consistency mechanism as constraints for making compatible predictions across these variants. We update PLMs by using parameter-efficient fine-tuning (PEFT), leading to more accurate predictions in few-shot and zero-shot fact verification tasks.
Our experiments on three public verification datasets show that ProToCo significantly outperforms state-of-the-art few-shot fact verification baselines.
With a small number of unlabeled instances, ProToCo also outperforms the strong zero-shot learner T0 on zero-shot verification. Compared to large PLMs using in-context learning (ICL) method, ProToCo outperforms OPT-30B and the Self-Consistency-enabled OPT-6.7B model in both few- and zero-shot settings.

\end{abstract}

\section{Introduction}

The problem of misinformation has sparked significant attention on the task of fact verification within the natural language processing (NLP) community. Such task, typically represented by Fact Extraction and VERification (FEVER) benchmark~\cite{thorne-etal-2018-fever}, requires models to verify if pieces of evidence \textit{support}, \textit{refute}, or contain \textit{not enough information (NEI)} to validate a given claim. 

Fully supervised fact verification has been widely studied and achieved good performance on the data of different domains~\citep{Nie_Chen_Bansal_2019,ma-etal-2019-sentence,wadden-etal-2020-fact,guo-etal-2022-survey}. However, collecting a large set of training data is labor-intensive, time-consuming and costly especially with the constant emergence of new events, such as COVID-19~\citep{lee-etal-2021-towards,pan-etal-2021-zero,saakyan-etal-2021-covid}, that may be out-of-domain. 
Few-shot fact verification is an urgent need but has been paid little attention because its performance is previously not competitive given very few training data~\cite{lee-etal-2021-towards,zeng2022aggregating}, not to mention the zero-shot setting without any labeled data available at all.

\begin{figure}[!t]
  \centering
  \includegraphics[width=\linewidth]{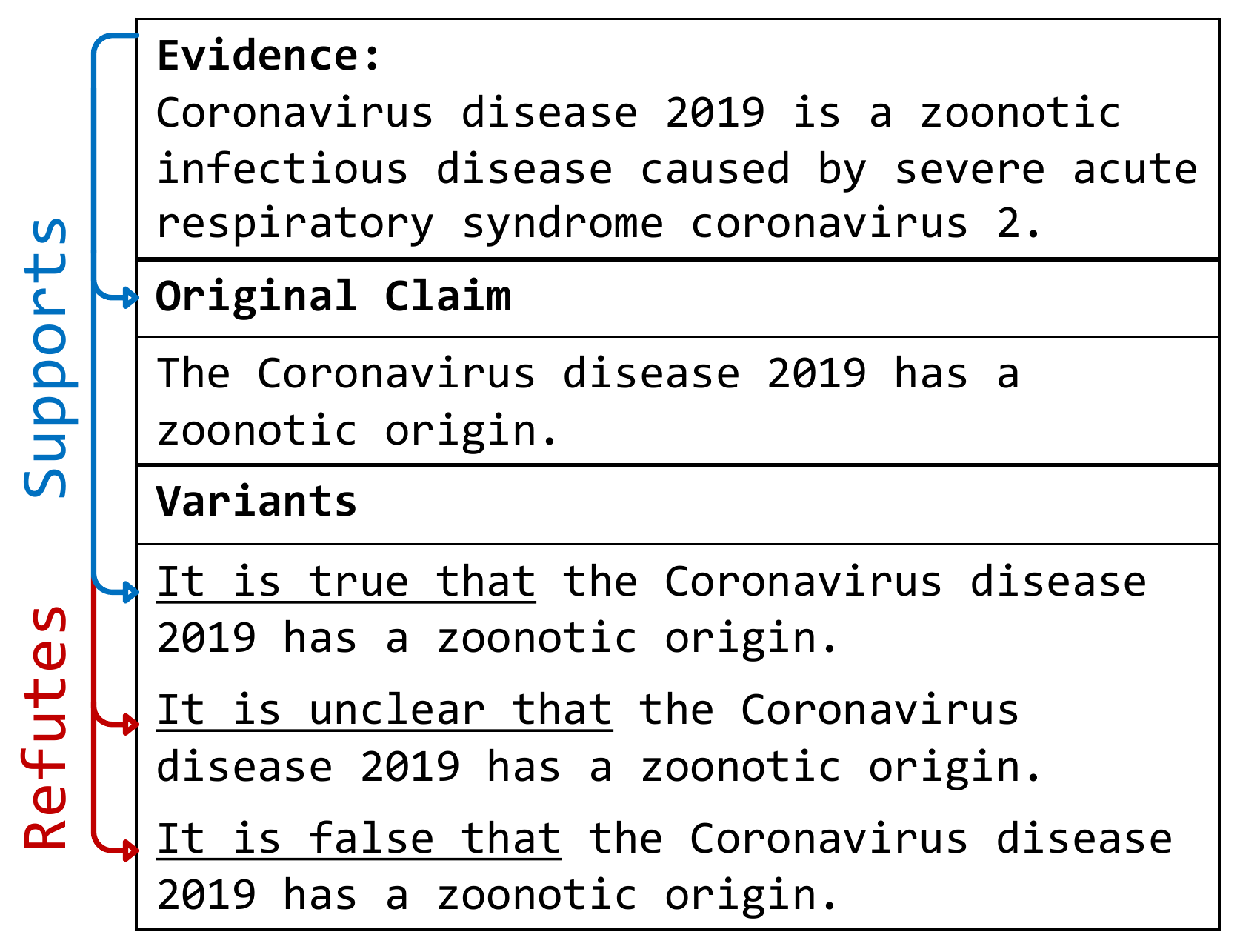}
  \caption{An illustration of our consistency mechanism for evidence-based fact verification when the evidence supports the claim. The PLM's judgements on the variants should be logically consistent across the different variants of the claim.}
  \label{fig:intro-consistency}
\end{figure}

In this paper, we try to improve PLMs' capability on factuality assessment for few-shot and zero-shot evidence-based fact verification. In general, consistency in fact verification dictates our assessment on the veracity of a claim based on the evidence given. For example, Figure~\ref{fig:intro-consistency} shows that given the same evidence and three major variants of the claim, the judgement of factuality on the \emph{confirmation variant} ``\texttt{It is true that [claim]}'' should remain the same as that of the original claim, while the judgement on the \emph{uncertainty variant} ``\texttt{It is unclear that [claim]}'' and \emph{negation variant} ``\texttt{It is false that [claim]}'' should be opposite to that of the original claim. The relations (i.e., confirmation, uncertainty and negation) between the claim and its variants naturally constrain what decisions should be made for the variants once the decision on the claim is determined, and vice versa. 
Such simple consistency constraints with minor adjustments can be generalized to different cases (e.g., when the evidence refutes the claim. See Section~\ref{sect:logco} for detail).
Meanwhile, prior studies on consistency in other domains (e.g., knowledge base and question answering (QA)) have shown a strong correlation between PLM's performance and their self consistency~\cite{elazar-etal-2021-measuring,wang2022self}, but it is empirically observed that PLMs are insufficient to transfer self-consistency to downstream tasks~\cite{ettinger-2020-bert,kassner-schutze-2020-negated,kassner-etal-2021-beliefbank,elazar-etal-2021-measuring}.
We therefore aim to explicitly impose consistency on PLMs for improving few-shot and zero-shot fact verification performance.  
 
Inspired by the recent success of prompt-enabled PLMs on various few-shot NLP tasks via forming natural language prompts using templates~\cite{radford2019language,brown2020language,gao-etal-2021-making,liu2022few}, we construct the variants of a given claim by simply altering prompt templates while keeping the claim itself unchanged. Further, we define a factuality-grounded consistency mechanism based on the aforementioned relations between the claim and its variants, and assign the labels (i.e., \textit{support}, \textit{refute}, and \textit{NEI}) satisfying the consistency to the variants, so that we obtain a set of claim-evidence pairs with consistency constraints. To bring such consistency to PLMs, we then use these pairs to fine-tune T-Few~\cite{liu2022few}, a prompt-enabled PLM with a parameter-efficient fine-tuning (PEFT) method by only updating a small number of parameters. 
We name our method as ProToCo, \underline{Pro}mpt PLMs \underline{To} be \underline{Co}nsistent, for improving the consistency of PLMs for few-shot and zero-shot fact verification. Our main contributions can be summarized as follows~\footnote{Code and dataset are available at \url{https://github.com/znhy1024/ProToCo}}:
\begin{itemize}[leftmargin=*]
    \item We design a general factuality-grounded consistency scheme to provide explicit consistency constraints for improving few-shot fact assessment, which is generalizable to zero-shot setting.  
    \item We propose ProToCo, a novel prompt-based consistency training method for improving PLMs on few-shot and zero-shot fact verification.
    \item Evaluation results on three public fact verification datasets from different domains confirm that ProToCo outperforms the state-of-the-art few-shot baselines by up to 30.4\% relative improvement in terms of F1, and also consistently outperforms the strong zero-shot learner T0-3B~\cite{sanh2022multitask} in zero-shot verification. 
    \item When compared to large PLMs in both settings, ProToCo achieves overall better performance than OPT-30B~\cite{zhang2022opt} and significantly outperforms the Self-Consistency-enabled OPT-6.7B model based on Chain-of-Thought (CoT) prompting~\cite{wang2022self}.
\end{itemize}

\section{Related Work}

Existing methods tried to address few-shot fact verification by utilizing the implicit knowledge of PLMs encoded in their parameters without gradients update. 
\citet{lee-etal-2021-towards} hypothesized that the perplexity of concatenated claim-evidence text sequence evaluated by a language model could benefit claim verification, and used a few training instances to find the threshold of perplexity scores for determining the label of test claim.
~\citet{zeng2022aggregating} utilizes PLMs to create a set of representative vectors for each class based on the semantic difference between claim and evidence of a few training instances, which are used to label test claims based on Euclidean distance during inference. 
However, these models do not update model parameters solely relying on the pre-encoded knowledge of PLMs, which cannot improve the language model itself and may not generalize well in new domains. And they also cannot perform zero-shot task as a few labeled instances are required as the anchors for labeling new instances.
Our method aims to update PLMs efficiently to utilize new knowledge in a few examples and enforce model's consistency for improving both few-shot and zero-shot verification.

Recently, several studies worked to generalize PLMs to the target domain by fine-tuning the model with the full training dataset of fact verification from a different domain~\cite{wadden-etal-2020-fact,saakyan-etal-2021-covid,schuster-etal-2021-get,wadden-etal-2022-multivers}. Meanwhile, some works targeted to instruct PLMs to generate task-specific training data used to fully train a classifier for fact verification~\cite{pan-etal-2021-zero,wright-etal-2022-generating}. Such works need a carefully crafted generation policy based on real corpus of the task, and the performance heavily depends on the quality of generated data. These approaches are considered distantly supervised, and significantly differ from ours as they do not aim to build any few-shot or zero-shot model. Unlike these studies, we assume that the language model is minimally aware of fact verification task with only a few task-specific examples, which may be even unlabeled.

In general, PLMs have shown strong few-shot learning ability in various NLP tasks~\cite{brown2020language,sanh2022multitask}. In-context learning (ICL) uses natural language prompts or instructions to elicit desired output from PLMs without gradient updates~\cite{radford2019language,brown2020language}. However, ICL is hard to deal with many prompted instances~\cite{liu2022few}, sensitive to the prompt design~\cite{liu-etal-2022-makes,lu-etal-2022-fantastically} and performs worse than fine-tuning~\cite{brown2020language,liu2022few}.
An alternative approach is parameter-efficient fine-tuning (PEFT) by updating only a small number of parameters to bridge the gap with standard fine-tuning~\cite{houlsby2019parameter,he2021towards,karimi2021compacter,lester-etal-2021-power,wei2021finetuned,ben-zaken-etal-2022-bitfit,liu2022few}. 
Our method utilizes T-few~\cite{liu2022few}, a state-of-the-art PEFT-enabled model, as our backbone to perform the factuality-grounded consistency training.

Previous works evaluate the self-consistency of PLMs by modifying the context of input sentences ~\cite{ettinger-2020-bert,kassner-schutze-2020-negated,ravichander-etal-2020-systematicity,elazar-etal-2021-measuring} and empirically show that PLMs are insufficient to transfer self-consistency to downstream tasks.
Some works in question answering (QA) prompt large PLMs (e.g., GPT-3~\cite{brown2020language}) to improve QA accuracy by strengthening the consistency of predicted answers. ~\citet{wang2022self} prompts PLM to generate multiple explanations and candidate answers and choose the answer that consistently occurs as the prediction. The Maieutic prompting~\cite{jung2022maieutic} designed for True-or-False commonsense QA, and ConCoRD~\cite{mitchell2022enhancing} designed specifically based on self-consistency benchmarks, both of which elicit PLMs to generate distributions for possible candidate answers, followed by a MaxSAT solver~\cite{Battiti2009} to infer the most probable answer by eliminating contradictory candidates. Both methods are based on different consistency definitions from ours and may not be suitable for the fact verification task.

\section{Problem Definition}
Let $\mathbf{C}= \{(x_i,y_i)\}$ be a fact verification dataset, containing training set $\mathbf{C}_{train}$ and test set $\mathbf{C}_{test}$, where each instance consists of the input $x_i$ and ground-truth label $y_i \in \mathcal{Y}$ and $\mathcal{Y}=\{\texttt{Support,  NEI, Refute}\}$. 
Let $x_i=(c_i,e_i)$, and the task aims to predict if the given pieces of evidence $e_i$ supports, refutes or has not enough information to validate the claim $c_i$. 
In the few-shot setting, we randomly sample $K$ instances \emph{per class} from $\mathbf{C}_{train}$ for training as the class distribution is unknown. As a result, the total number of instances is $3K$ and the few-shot training set is denoted as $\mathbf{C}_{train}^{fs} = \{(x_i,y_i)\}^{3K}$.  
The zero-shot setting is similar but only uses $x_i$ for each instance and the \emph{unlabeled} training set is given as $\mathbf{C}_{train}^{zs} = \{(x_i)\}^{3K}$. Note that the absence of ground-truth label makes the setting zero-shot~\cite{wright-etal-2022-generating,Prompt-Consistency}. Similar to previous works~\cite{lee-etal-2021-towards,liu2022few}, we do not assume the availability of development set as it is more realistic in a limit data scenario.
Our goal is to generalise a PLM $\mathcal{M}_\theta$ to the unseen test set $\mathbf{C}_{test}$, fine-tuned only using $\mathbf{C}_{train}^{fs}$ or $\mathbf{C}_{train}^{zs}$, where $\theta$ denotes language model parameters.

\section{Methodology}
\begin{figure*}[t]
  \centering
  \includegraphics[width=\textwidth]{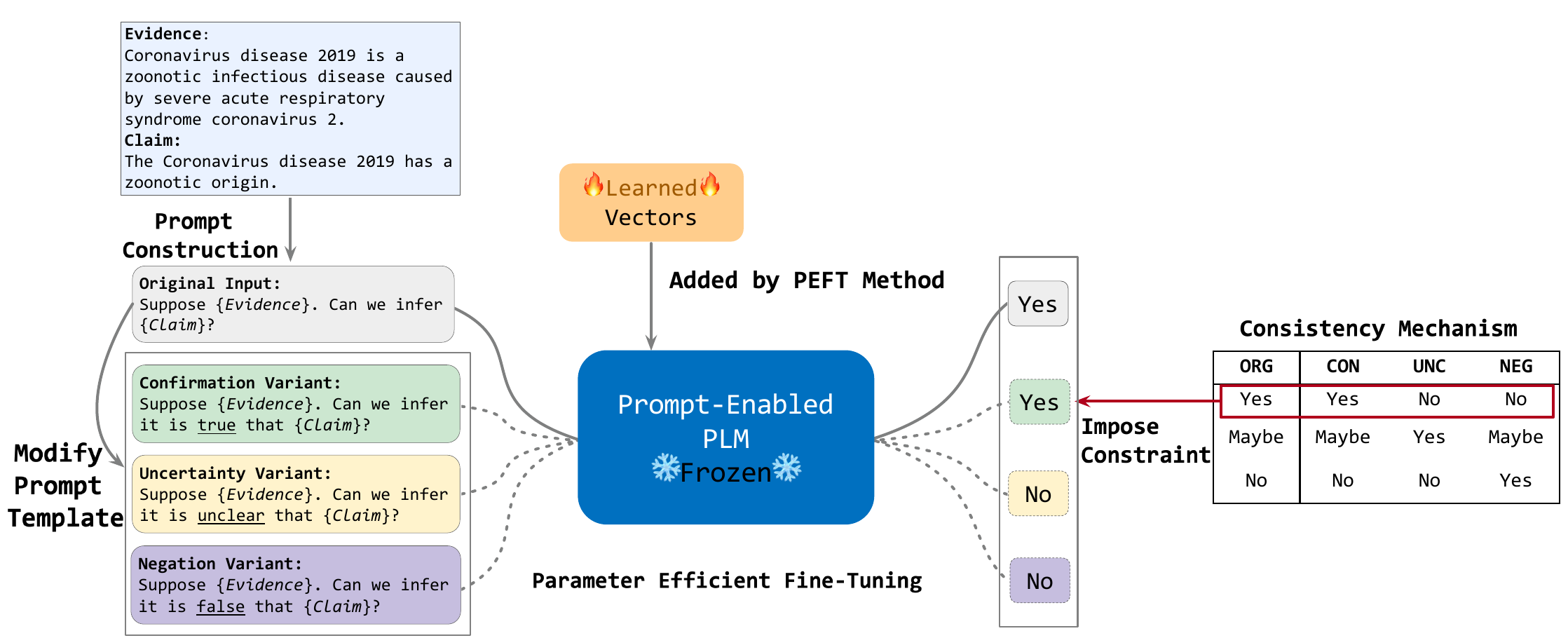}
  \caption{The architecture of our ProToCo model. Given a claim-evidence pair, confirmation variant (CON), uncertainty variant (UNC) and negation variant (NEG) are created by modifying the prompt template. The original input (ORG) and its variants are used to train the PLM. We use a PEFT method (i.e., $(\text{IA})^3$) to train the PLM, which only updates the parameters of additionally learned vectors while other parameters are frozen. Consistency will be imposed as the constraints on PLM's predictions over the claim and its variants.}
  \label{fig:model}
\end{figure*}

\subsection{Prompt Construction} \label{prompting}
Given a labelled instance $(x_i,y_i)$, the input $x_i$ (i.e., $c_i$ and $e_i$) and the label $y_i$ are firstly reformatted as a natural language input and response using a prompt template $\mathcal{T}$, which consists of an input template $\mathcal{T}_x$ and a target template $\mathcal{T}_y$. For example, as shown in Figure \ref{fig:model}, the reformatted input $\mathcal{T}_x(x_i)$ is obtained by filling the evidence and claim in their corresponding fields:
\begin{align*}
    \texttt{Suppose}\; \{e_i\}.\texttt{ Can we infer }\{c_{i}\}\texttt{?}
\end{align*}
and the reformatted label can be $\mathcal{T}_y(y_i)$ = \texttt{Choices[$y_i$]}. Here \texttt{Choices} is a prompt-specific target words mapping containing response keys \texttt{\{Yes, Maybe, No\}}, where \texttt{Yes} is mapped to \texttt{Support}, \texttt{Maybe} to \texttt{NEI}, and \texttt{No} to \texttt{Refute}.

\subsection{Inference} \label{infer}

We take the text-to-text PLM (e.g., T5 ~\cite{raffel2020exploring}) as $\mathcal{M}_\theta$ since the prompted input and output are text sequences. Let $\mathcal{V}$ be the vocabulary of $\mathcal{M}_\theta$. We denote each $\mathcal{T}_x(x_i)$ as an input sequence of tokens $\mathbf{x}_i$ and $\mathcal{T}_y(y_i)$ as a target sequence of tokens $\mathbf{y}_i=\{t_j \in \mathcal{V}, j\in[1,|\mathbf{y}_i|]\}$ to be generated. Then, the probability of the target sequence is $p_\theta(\mathbf{y}_i \mid \mathbf{x}_i) = \prod_{j=1}^{|\mathbf{y}_i|} p_\theta(t_j \mid \mathbf{x}_i,t_{<j})$, where $p_\theta(t_j \mid \mathbf{x}_i,t_{<j})$ is the probability of each token $t_j$ assigned by the model $\mathcal{M}_\theta$ during autoregressive generation given the input sequence $\mathbf{x}_i$ and the tokens generated prior to $t_j$. 
Since the sequence $\mathbf{y}_i$ corresponds to the class $y_i$, the predicted score for class $y_i$ given by $\mathcal{M}_\theta$ can be defined as the log-probability normalized by the length of output sequence to avoid possible bias towards length~\cite{liu2022few}:
\begin{equation}\label{eq:beta}
\beta(x_i,y_i,\mathcal{T})=\frac{1}{|\mathbf{y}_i|} \log p_\theta(\mathbf{y}_i \mid \mathbf{x}_i)
\end{equation}
In this way, we obtain the predicted scores of all classes using Equation~\ref{eq:beta} and use rank classification for inference by following~\cite{liu2022few}. All classes are ranked by the predicted scores and the top-ranked class is taken as the prediction.

\subsection{The Consistency Mechanism}\label{sect:logco}
In this section, we describe how to establish the consistency for fact verification task. 
Our goal is two-fold: 1) construct a set of variants for a claim corresponding to three basic logical relations between the claim and a variant, i.e, confirmation, uncertainty, and negation; 2) the labels of variants can be unambiguously derived based on the relations above once the label of original claim-evidence pair is known. To this end, we construct the logical variants by modifying the prompt input template $\mathcal{T}_x$, as shown in Figure \ref{fig:model}. 

Specifically, we prepend ``\underline{\texttt{it is \{$w$\} that}}'' before $c_i$ to get a claim's logical variants, where $w\in \mathcal{V}$ can be an affirmative word (e.g., true), an uncertain word (e.g., unclear), or a negative word (e.g., false), corresponding to the aforementioned relations.  Figure \ref{fig:model} shows the consistency constraints that the model should strive to satisfy based on the set of labels assigned to the original claim and its variants given $\mathcal{T}_y(y_i)$. For example, when $\mathcal{T}_y(y_i)$ is \texttt{Yes} or \texttt{No}, the label of the confirmation variant should be same as that of the original claim since they entail each other, while the negation variant should have the opposite label because of their contradiction, and the uncertainty variant is assigned as \texttt{No} since the evidence indicates it is sufficient to draw a certain conclusion. Situation is slightly different when $\mathcal{T}_y(y_i)$ is \texttt{Maybe} since there is not enough evidence to support or refute the confirmation and negation variants, and as a result, both confirmation and negation variants are designated as \texttt{Maybe} while the uncertainty variant is assigned as \texttt{Yes}.
With these consistency constraints, we could label the claim variants for each training instance and utilize them for fine-tuning the model.

\subsection{Training Strategy}

It is challenging for few-shot fine-tuning of PLMs as updating a large number of parameters with a few instances may result in unstable performance. Also, there are no labeled instances available for zero-shot fine-tuning.
We will introduce how to bring our consistency mechanism into the training of PLMs in both settings.

\subsubsection{Parameter Efficient Fine-Tuning (PEFT)}
The traditional fine-tuning methods updating all parameters of PLMs are found unstable in the few-shot setting~\cite{revisit-bert-finetuning,Mosbach-Stability,Dodge-ft-early}, and could be computationally expensive.  We thus employ PEFT methods~\cite{houlsby2019parameter,karimi2021compacter,lester-etal-2021-power,hu2021lora,ben-zaken-etal-2022-bitfit,liu2022few} for more efficient fine-tuning. 

We exploit the T-Few recipe which applies a PEFT method called $\textsc{(IA)}^3$~\citep{liu2022few} on a zero-shot learner T0~\cite{sanh2022multitask} to enable its few-shot ability. The $\textsc{(IA)}^3$ modifies Transformer~\cite{vaswani2017attention} via multiplying the keys and values in attention and the intermediate activations of position-wise feed-forward networks by the learned vectors, so that a small number of parameters are introduced for fine-tuning. And T0 has been endowed with a strong zero-shot generalizability by training a LM-adapted T5~\cite{lester-etal-2021-power} on a set of datasets covering numerous NLP tasks, where each training instance is reformatted as a natural language input and response using a prompt template.

\subsubsection{Loss Functions}
With a few training instances, we follow \citet{liu2022few} by combining several different loss functions to update the new parameters.

\begin{itemize}[leftmargin=*]
\item \textbf{Standard cross-entropy loss} encourages $\mathcal{M}_\theta$ to assign higher probability $ p_\theta(\mathbf{y}_i \mid \mathbf{x}_i)$ to the correct target sequence $\mathbf{y}_i$ given the input sequence $\mathbf{x}_i$:
\begin{equation}
     \mathcal{L}_i^{\mathrm{lm}} = -\frac{1}{|\mathbf{y}_i|} \log p_\theta(\mathbf{y}_i \mid \mathbf{x}_i)
\end{equation}

\item \textbf{Classification task loss} is based on cross-entropy. Given the predicted scores $\beta(x_i,y_i,\mathcal{T})$ assigned by PLM, the probability of predicting class $y_i$ can be calculated as:
\begin{equation}
     q_\theta(y_i \mid x_i) =\frac{\exp{(\beta(x_i,y_i,\mathcal{T}))}}{\sum_{y'\in \mathcal{Y}}\exp{(\beta(x_i,y',\mathcal{T}))}}
\end{equation}
and the loss for the task is:
\begin{equation}
     \mathcal{L}_i^{\mathrm{cls}} = -\log q_\theta(y_i \mid x_i)
\end{equation}

\item \textbf{Unlikelihood loss} forces incorrect target sequences to be assigned with lower probabilities~\cite{welleck2020neural}:
\begin{equation}
\small
 \mathcal{L}_i^{\mathrm{ul}}=-\frac{\sum_{k=1,k\ne i}^{|\mathcal{Y}|-1} \sum_{j=1}^{|\mathbf{y}_k|} \log (1-p_\theta(t_{j} \mid \mathbf{x}_i, t_{<j}))}{\sum_{k=1,k\ne i}^{|\mathcal{Y}|-1}|\mathbf{y}_k|}
\end{equation}
\end{itemize}

The total loss for fine-tuning our backbone model T-Few is a sum of the above three losses: $\mathcal{L}=\sum_{i=1}^{3K}\mathcal{L}_i^{\mathrm{lm}}+\mathcal{L}_i^{\mathrm{cls}}+\mathcal{L}_i^{\mathrm{ul}}$.

\subsubsection{Few-Shot and Zero-Shot Training}

In the few-shot setting, we first fine-tune the model with the original labeled instances as a warm-up, and then continue the fine-tuning with the created variants and the logically consistent labels which are derived from the claim following the proposed consistency mechanism (see Section~\ref{sect:logco}).

Given no labeled instance in the zero-shot setting, we directly fine-tune the model with the variants using the following strategy: at each training step, the prediction of the original instance by the PLM is used to assign pseudo labels to its variants based on the proposed consistency mechanism. To some extent, this training strategy provides a regulation to PLM and guide it to update the prediction on the original instance.
Note that such method is still zero-shot since what is considered in training is only the determined logical relations between the claim and its variants and no ground-truth information is exploited.

\section{Experiments and Results}

\subsection{Experimental Setup}

\begin{table}[t!]
\centering
\small
\normalsize
\begin{adjustbox}{width={\linewidth},totalheight={!},keepaspectratio}%
\begin{tabular}{ccccc}
\toprule[1.0pt]
Dataset & label & Supports  & Refutes   & NEI\\ 
\midrule[0.5pt]
\multirow{2}{*}{FEVER} 
&Train  & 80,035          & 29,775   & 35,639\\
&Test  & 3,333          & 3,333   & 3,333\\
\midrule[0.5pt]
\multirow{2}{*}{SciFACT}  
&Train  & 332          & 173   & 304\\
&Test  & 124          & 64   & 112\\
\midrule[0.5pt]
\multirow{2}{*}{VitaminC}  
&Train  & 124,864          & 71,108   & 52,981\\
&Test  & 17,306          & 9,907   & 7,268\\
\bottomrule[1.0pt]
\end{tabular}
\end{adjustbox}
\caption{Statistics of three datasets used for evaluation. 
}
\label{tab:dataset statistics}
\end{table}

\subsubsection{Datasets}
We use three public fact verification datasets from different domains. Their statistics are shown in Table \ref{tab:dataset statistics}.
\textbf{FEVER}~\cite{thorne-etal-2018-fever} provides manually crafted claims by altering factual sentences from Wikipedia. The claims are classified as \texttt{Support}, \texttt{Refute} or \texttt{NEI} by annotators. This dataset only provides gold evidence for the \texttt{Support} and \texttt{Refute} classes. 
To provide evidence for instances in the \texttt{NEI} class, we randomly sample a sentence from Wikipedia for each claim using uniform sampling method by following~\citet{thorne-etal-2018-fever}. 
\textbf{SciFACT}~\cite{wadden-etal-2020-fact} is a fact verification dataset which consists of expert-written scientific claims by re-writing citation sentences occurring in biomedical literature. We choose the sentence from the cited abstract with the highest TF-IDF similarity to the claim for the \texttt{NEI} class following~\citet{wadden-etal-2020-fact}.
\textbf{VitaminC}~\cite{schuster-etal-2021-get} is a challenging dataset with cases requiring models to identify subtle factual changes. It is created by utilizing Wikipedia revisions that alter a factual statement to create claim-evidence pairs, where the instances for each revision are made contrastive, i.e., they contain evidence pairs that are nearly identical in content, but one supports the claim while the other contradicts it. VitaminC has three classes similar to FEVER and provides real or synthetic revisions. We only use instances from real revisions as the synthetic does not include the \texttt{NEI} class.

\subsubsection{Baselines}
We compare ProToCo to the following few-shot baselines: \textbf{Majority} simply assigns the most frequent class of the training set to all instances;
\textbf{RoBERTa-L}~\cite{liu2019roberta} is a pre-trained RoBERTa-large model with a feed-forward classifier fine-tuned on top of it; 
\textbf{GPT2-PPL}~\cite{lee-etal-2021-towards} uses a few labeled instances to find the threshold of perplexity scores based on the GPT-2 language model~\cite{radford2019language} for determining claim class labels; \textbf{SEED}~\cite{zeng2022aggregating} utilizes PLMs to obtain semantic difference vectors between claims and their evidence and average them to create representative vectors for each class, which are used to label instances based on Euclidean distance during inference. 

We also compare ProToCo to zero-shot baselines: \textbf{T0}~\cite{sanh2022multitask} is a strong zero-shot learner which is created by training LM-adapted T5~\cite{lester-etal-2021-power} on datasets covering multiple tasks, where each training instance is converted as prompted input and output; \textbf{T-Few}~\cite{liu2022few} additionally pre-trains the new parameters introduced by $(\textsc{IA})^3$ based on T0.

\begin{table*}[t!]
\small
\begin{adjustbox}{width={\linewidth},keepaspectratio}%
\begin{tabular}{ccccccccc|ccc}
\toprule[1.0pt]
\multirow{2}{*}{Datasets}  &
\multicolumn{8}{c}{Few-shot Methods} & \multicolumn{3}{|c}{Zero-shot Methods}\\
\cmidrule{2-12} 
      & Majority & RoBERTa-L & $\text{GPT2-PPL}_\text{base}$   & $\text{GPT2-PPL}_\text{xl}$      & $\text{SEED}_\text{nli}$   & $\text{SEED}_\text{mpnet}$     & T-Few & ProToCo  
      & T0-3B  & T-Few   & ProToCo\\ 
\midrule[0.5pt]
SciFACT
                      & 0.195 & \makecell{0.210 \\ \small{(0.09)} \\ \addlinespace[0.1cm] } & \makecell{0.326 \\ \small{(0.04)}\\ \addlinespace[0.1cm] }  &\makecell{0.348 \\ \small{(0.06)} \\ \addlinespace[0.1cm] }   & \makecell{0.355 \\ \small{(0.05)}\\ \addlinespace[0.1cm] } &\makecell{0.273 \\ \small{(0.07)} \\ \addlinespace[0.1cm] } & \makecell{\underline{0.382} \\ \small{(0.05)}\\ \addlinespace[0.1cm] } & \makecell{\textbf{0.498} \\ \small{(0.03)}\\ \addlinespace[0.1cm] }
                      & \makecell{\underline{0.315} \\ \small{(0.05)}\\ \addlinespace[0.1cm] }          & \makecell{0.305 \\ \small{(0.03)}\\ \addlinespace[0.1cm] }   & \makecell{\textbf{0.331} \\ \small{(0.02)}\\ \addlinespace[0.1cm]}\\                
\midrule[0.5pt]
FEVER
                   & 0.167 & \makecell{0.169 \\ \small{(0.01)} \\ \addlinespace[0.1cm] } & \makecell{0.293 \\ \small{(0.04)}\\ \addlinespace[0.1cm] }  &\makecell{0.329 \\ \small{(0.10)} \\ \addlinespace[0.1cm] }   & \makecell{0.501 \\ \small{(0.07)}\\ \addlinespace[0.1cm] }  &\makecell{0.352 \\ \small{(0.02)} \\ \addlinespace[0.1cm] }  & \makecell{\underline{0.851} \\ \small{(0.06)}\\ \addlinespace[0.1cm] } & \makecell{\textbf{0.891} \\ \small{(0.03)}\\ \addlinespace[0.1cm] }& \makecell{\underline{0.446} \\ \small{(0.03)}\\ \addlinespace[0.1cm] }        & \makecell{0.433 \\ \small{(0.01)}\\ \addlinespace[0.1cm] }  & \makecell{\textbf{0.479} \\ \small{(0.00)}\\ \addlinespace[0.1cm] } \\
\midrule[0.5pt]
VitaminC
                   & 0.223 & \makecell{0.146 \\ \small{(0.02)} \\ \addlinespace[0.1cm] } & \makecell{0.303 \\ \small{(0.04)}\\ \addlinespace[0.1cm] }  &\makecell{0.327 \\ \small{(0.04)} \\ \addlinespace[0.1cm] }  & \makecell{0.313 \\ \small{(0.05)}\\ \addlinespace[0.1cm] } &\makecell{0.306 \\ \small{(0.04)} \\ \addlinespace[0.1cm] } & \makecell{\underline{0.489} \\ \small{(0.09)}\\ \addlinespace[0.1cm] } & \makecell{\textbf{0.520} \\ \small{(0.05)}\\ \addlinespace[0.1cm] }& \makecell{0.373 \\ \small{(0.02)}\\ \addlinespace[0.1cm] }          & \makecell{\textbf{0.400} \\ \small{(0.00)}\\ \addlinespace[0.1cm] } & \makecell{\underline{0.386} \\ \small{(0.00)}\\ \addlinespace[0.1cm] } \\
                  
\bottomrule[1.0pt]
\end{tabular}
\end{adjustbox}
\caption{Results of different few-/zero-shot fact verification methods in 4-shot and 0-shot settings on three datasets. We report the macro-F1 averaged over 4 trials with randomly selected training samples from the datasets using different seeds. The best results are in bold while the second results are underlined. The standard deviation is in (.).}
\label{tab:experiments}
\end{table*}

\subsubsection{Experimental Settings}
For few-shot fact verification, we report 4-shot experiments as the main result. We also conduct $K$-shot experiments for $K=\{1,2,4,8,16\}$ reported as supplementary results. 
For zero-shot experiments, we randomly sample 30 instances per class from each training set for fine-tuning. Note that no labels are used in this setting.
For fair and robust comparison, we sample the training instances based on four random seeds and report the mean performance of macro-F1 and standard deviation over these four splits in all experiments. The seeds and data splits are kept the same across different models.

We use the original source code\footnote{\url{https://github.com/r-three/t-few}} of T-Few~\cite{liu2022few} with its released pre-trained checkpoint of 3B parameters as our backbone model. Following the T-Few paper, we randomly sample a prompt template from the Public Pool of Prompts (P3)~\cite{bach-etal-2022-promptsource} for each instance at \emph{each} training and inference step to increase the diversity and variability of prompts used.
We set training steps as 1,500, batch size as 4, and learning rate as \num{1e-4} for both few-shot and zero-shot settings\footnote{For all methods, we use the number of shots as batch size if the training size is less than the batch size.}.

For fine-tuning the RoBERTa-L model, we follow~\citet{lee-etal-2021-towards} using \num{2e-5} as learning rate and 32 as batch size, and train it for 10 epochs. We use the original code of GPT2-PPL\footnote{\url{https://github.com/HLTCHKUST/Perplexity-FactChecking}} and conduct experiments using GPT2-base as the backbone following the original setting. Additionally, we also present the results of GPT2-PPL with a larger backbone GPT2-xl\footnote{To deal with 3-way classification, we separate support and unsupported classes first, and then separate NEI and refutes classes from the predicted unsupported class, following the assumption that misinformation has higher perplexity~\cite{lee-etal-2021-towards}}. 
We reproduce SEED following the original implementation details in the paper~\cite{zeng2022aggregating} with $\text{BERT}_{\text{nli}}$\footnote{\url{https://huggingface.co/sentence-transformers/bert-base-nli-mean-tokens}} as its base model which was fine-tuned on NLI tasks. Furthermore, we report the results of SEED using the pre-trained model all-mpnet-base-v2\footnote{\url{https://huggingface.co/sentence-transformers/all-mpnet-base-v2}} as backbone since it provides the best quality of sentence embeddings in all pre-trained models of sentence transformers~\cite{reimers-gurevych-2019-sentence}
\footnote{\url{https://www.sbert.net/docs/pretrained_models.html\#model-overview}}. 
We use the code and pre-trained checkponit with 3B parameters of T0 from Hugging Face Transformers\footnote{\url{https://huggingface.co/bigscience/T0}}.
All the experiments use a server with 4 NVIDIA Tesla-V100 32GB GPUs.

\begin{figure*}
    \centering
    \includegraphics[width=\linewidth]{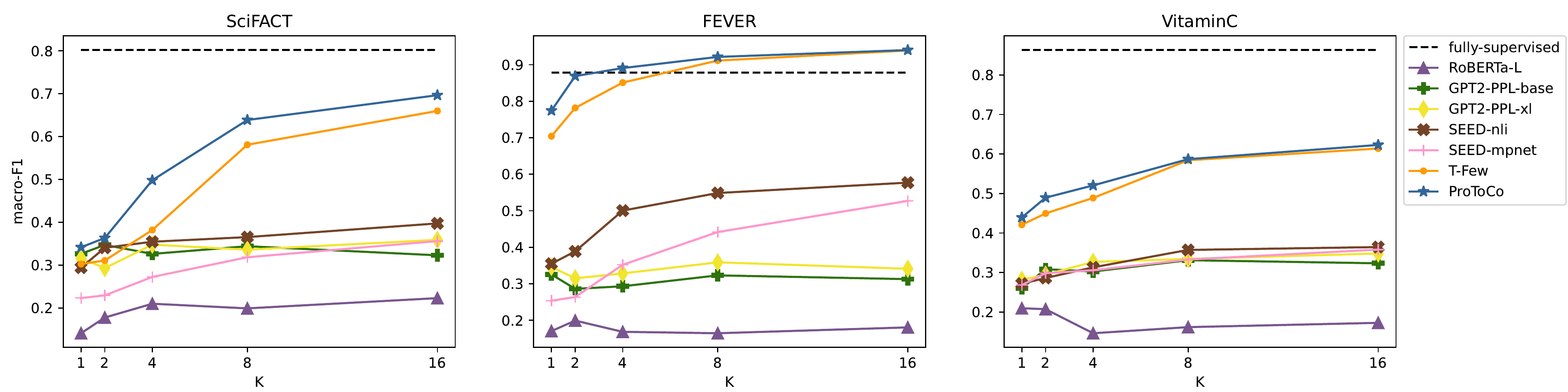}
    \caption{The performance comparison under different the number of shot $K$. For all $K$ tested, ProToCo consistently outperforms all the baselines. We report the results of fully-supervised models using oracle evidence as a reference: the results of RoBERTa-large model from~\citet{pradeep-etal-2021-scientific} and~\citet{pan-etal-2021-zero} on SciFACT and FEVER, respectively; the result of ALBERT-xlarge model~\cite{lan-etal-2020-learning} on VitaminC is obtained by evaluating test set using the provided checkpoint and original code from~\citet{schuster-etal-2021-get}.}
    \label{fig:different k}
\end{figure*}

\subsection{Few-Shot Result} \label{Experiments Results}
The results of few-shot fact verification are reported in Table~\ref{tab:experiments}. We have the following observations.

\textbf{Firstly}, given very few labeled instances, RoBERTa-L does not always improve few-shot performance, which is consistent with the empirical finding that traditional fine-tuning of PLMs is unstable in the few-shot setting~\cite{revisit-bert-finetuning,Mosbach-Stability,Dodge-ft-early}. 

\textbf{Secondly}, with the designs for few-shot learning on PLMs, both versions of GPT2-PPL and SEED achieve much better performance than the majority class and RoBERTa-L, without any gradient update. With different backbone models, $\text{GPT2-PPL}_\text{xl}$ outperforms $\text{GPT2-PPL}_\text{base}$ due to its larger model size, while $\text{SEED}_\text{mpnet}$ lags far behind $\text{SEED}_\text{nli}$ possibly because the base model $\text{BERT}_{\text{nli}}$ fine-tuned on NLI task can be more readily adapted to the fact verification task compared to the base model all-mpnet-base-v2, which was fine-tuned on the sentence matching task.
On SciFACT and FEVER datasets, $\text{SEED}_\text{nli}$ with the semantic difference vector outperforms $\text{GPT2-PPL}_\text{xl}$ that predicts labels based on a perplexity score. 
However, $\text{SEED}_\text{nli}$ is less advantageous on VitaminC as the semantic vector becomes less likely to identify the subtle factual differences in the contrastive instances. 

\textbf{Thirdly}, our backbone model T-Few clearly outperforms both versions of GPT2-PPL and SEED, which indicates that only relying on the implicit knowledge of PLMs without parameter update is insufficient for few-shot fact verification. Also, compared to RoBERTa-L, the obtained improvements on all datasets shows the PEFT method $(\text{IA})^3$ helps address the instability issue of traditional fine-tuning methods on PLMs under few-shot setting.

\textbf{Lastly}, ProToCo with consistency training leads to consistent gains on all datasets, considerably improving T-Few by 30.4\%, 6.3\% and 4.7\% on SciFACT, VitaminC and FEVER, respectively, which demonstrates the effectiveness of imposing the consistency constraints on model training. 

\subsection{Zero-Shot Result}
We examine the effectiveness of ProToCo in zero-shot setting, where it only uses a small number of \emph{unlabelled} instances for training. The zero-shot result is also given in Table~\ref{tab:experiments}. 

We can see that ProToCo performs better than T0-3B on all the datasets, achieving improvements by 7.4\%, 5.1\% and 3.5\% F1 on FEVER, SciFACT and VitaminC, respectively. And our consistency training also improves T-Few by 10.6\% and 8.5\% in FEVER and SciFACT, respectively. However, ProToCo performs slightly worse than T-Few on VitaminC. Given the contrastive construction approach of VitaminC dataset, we conjecture that this is might be because consistency training alone may not be able to effectively enhance the base model's ability to distinguish the contrastive instances without any supervision signals or prior adversarial training for the base model. One possible solution to address the issue is to use a stronger base model, which is pre-trained with adversarial data, to better capture the subtle differences in the contrastive instances. We will leave this to future work.

\subsection{Impact of Shots Number}
Figure \ref{fig:different k} illustrates the comparison between few-shot baselines and ProToCo as the number of shots $K$ increases. ProToCo consistently outperforms the few-shot baselines at all $K$ on the three datasets. 
The curves of both versions of SEED and GPT2-PPL models quickly saturate compared to ProToCo and changing to a larger backbone cannot bring much improvements in GPT2-PPL method as $K$ increases, suggesting that fine-tuning PLMs is necessary for improving few-shot performance for new knowledge to be learnt.

Interestingly, the improvement of ProToCo over T-Few becomes clearly smaller as $K$ increases on FEVER and VitaminC that are based on Wikipedia data (as Wikipedia-like data might be seen during PLM pre-training), but on the scientific domain dataset SciFACT, consistency training still can lead to a modest improvement even when $K$ reaches 16 shots and is inclined to grow continuously. This indicates the consistency training is especially helpful when the PLMs knows little about the type of data in Scientific domain.

As $K$ increases, ProToCo continues to narrow the gap with the fully-supervised model that was fine-tuned on the full training set. On FEVER, only using 4 labeled instances per class, it already outperforms the fully-supervised model. On VitaminC, however, the trend suggests that its performance is not likely to catch up with the fully-supervised model. Our analysis shows that the chance is low to be able to sample contrastive instances into such limited number of shots of training data. As a consequence, the contrastive nature of this dataset might be underrepresented by the sampled instances, potentially limiting the model from effectively learning such features. We believe that using more training instances or a base model pre-trained on contrastive data might boost ProToCo's performance on VitaminC, but we will leave this to future work.

\begin{table}[t]
    \centering
    \small
    \begin{tabular}{c|cc|cc}
    \toprule[1.0pt]
    \multirow{3}{*}{Dataset} &
 \multicolumn{2}{c}{ICL (OPT-30B)}  & \multicolumn{2}{|c}{ProToCo (T-Few)}\\ 
    \cmidrule{2-5}
    & zero-shot & 1-shot & zero-shot & 1-shot \\
    
    \midrule[0.5pt]
    SciFACT           & \makecell{0.332 \\ - \\ \addlinespace[0.1cm] }   & \makecell{0.324 \\ \small{(0.08)} \\ \addlinespace[0.1cm] }
    &\makecell{0.331 \\ \small{(0.02)} \\ \addlinespace[0.1cm] }
    &\makecell{0.342 \\ \small{(0.05)} \\ \addlinespace[0.1cm] }\\
    \midrule[0.5pt]
    FEVER           & \makecell{0.347 \\ \small{-} \\ \addlinespace[0.1cm] }  & \makecell{0.442 \\ \small{(0.03)} \\ \addlinespace[0.1cm] } & \makecell{0.479 \\ \small{(0.00)} \\ \addlinespace[0.1cm] }  & \makecell{0.774 \\ \small{(0.04)} \\ \addlinespace[0.1cm] }\\
    \midrule[0.5pt]
    VitaminC         & \makecell{0.340 \\ \small{-} \\ \addlinespace[0.1cm] } & \makecell{0.284 \\ \small{(0.08)} \\ \addlinespace[0.1cm] }& \makecell{0.386 \\ \small{(0.00)} \\ \addlinespace[0.1cm] }& \makecell{0.439 \\ \small{(0.04)} \\ \addlinespace[0.1cm] }\\
    \bottomrule[1.0pt]
    \end{tabular}
    \caption{Comparison between ProToCo (T-Few) and ICL (OPT-30B). Only an instruction is provided to OPT-30B in zero-shot setting. In few-shot setting, both task instruction and 3 training instances (1 shot) are provided.}
    \label{tab:opt30}
\end{table}

\subsection{Comparison to ICL of Large PLMs}
We compare ProToCo to ICL of relatively large PLMs in both few-shot and zero-shot settings. Specifically, we compare to OPT~\cite{zhang2022opt} with 30B parameters\footnote{\url{https://huggingface.co/docs/transformers/model_doc/opt}} -- 10 times larger than ProToCo, which is an open-source large causal language model with similar performance as GPT-3~\cite{brown2020language}. Results in Table \ref{tab:opt30} show that ProToCo achieves much higher F1 score compared to the few-shot ICL with OPT-30B. Compared to the zero-shot ICL with OPT-30B, ProToCo clearly outperforms ICL on FEVER and VitaminC datasets and performs equally well on SciFACT. This confirms the effectiveness of ProToCo in both settings and demonstrates how the consistency training method enables a smaller PLM to compete with the ICL method using a much larger PLM on fact verification task.

\begin{table}[t]
    \centering
    \small
      \begin{tabular}{cccc}
    \toprule[1.0pt]
    \multirow{2}{*}{Method} & \multicolumn{3}{c}{Datasets}\\
    \cmidrule{2-4}
    &SciFACT &FEVER &VitaminC\\ 
    \midrule[0.5pt]
    \makecell{SelfconCoT\\ \small(OPT-6.7B)}            & \makecell{0.289 \\ \small{(0.04)} \\ \addlinespace[0.1cm] }  &\makecell{0.358 \\ \small{(0.07)} \\ \addlinespace[0.1cm] } &\makecell{0.258 \\ \small{(0.06)} \\ \addlinespace[0.1cm] }\\
    \midrule[0.5pt]
    \makecell{ProToCo\\ \small(T-Few)}          &\makecell{0.342 \\ \small{(0.05)} \\ \addlinespace[0.1cm] } & \makecell{0.774 \\ \small{(0.04)}\\ \addlinespace[0.1cm]  }&\makecell{0.439 \\ \small{(0.04)}\\ \addlinespace[0.1cm] } \\
    \bottomrule[1.0pt]
    \end{tabular}
    \caption{Comparison with Self-Consistency Chain-of-Thought~\cite{wang2022self} using 3 training instances. The evaluation of FEVER and VitaminC are based on a random subset of test set given limited resources.}
\label{tab:self-con-cot}
\end{table}

\subsection{Comparison to Self-Consistency Models}
We compare ProToCo with the Self-Consistency Chain-of-Thought (SelfconCoT) method~\cite{wang2022self}, which samples multiple outputs from a language model and returns the most consistent answer in the set. We implement the SelfconCoT method following the details described in~\cite{wang2022self} and use OPT with 6.7B parameters as its base model and sample 20 outputs for each instance\footnote{Given the high cost of GPT-3.5 API and unavailability of checkpoints of PaLM~\cite{PaLM}, we have opted to utilize OPT as the base model and chosen the largest checkpoint OPT-6.7B that can be accommodated with our compute resources.}. Experiments are conducted with 3 training instances (1-shot) and evaluated on the full test set of SciFACT, and a random subset of the test set in FEVER and VitaminC given the limited resources we have. 

Results in Table~\ref{tab:self-con-cot} show that ProToCo significantly outperforms SelfconCoT on all datasets, despite the fact that the latter has 2 times more parameters, suggesting that PLM with our consistency training is more suitable for fact verification task. Additionally, using the same hardware, ProToCo is considerably more efficient than SelfconCoT, taking around 6 hours to finish 4 runs of experiments thanks to the PEFT method, while SelfconCoT needs around 14 hours.

\section{Conclusion and Future Work}
We propose a model called ProToCo to improve few- and zero-shot fact verification based on consistency training of PLMs. Experiments on three public datasets show that ProToCo achieves promising fact verification performance by outperforming the existing few- and zero-shot baselines, the in-context learning on large PLMs, and the self-consistency chain-of-thought method. Our method also outperforms fully-supervised model on FEVER dataset. 

In the future, we will explore few- and zero-shot solutions for other stages of fact-checking, e.g., evidence retrieval and justification generation, and combine them with ProToCo. We also plan to conduct experiments to evaluate the performance and level of consistency of larger language models (e.g., GPT-3~\cite{brown2020language}, InstructGPT~\cite{InstructGPT} and LLaMA~\cite{LLaMA}) on the fact verification task, when the computing resources are available.

\section{Limitations}
While ProToCo works well with our consistency training for improving fact verification under few-shot and zero-shot settings, our work has some limitations. Due to limited resources, currently we were unable to conduct comparison with larger PLMs and examine if extremely large models have already developed the similar or better level of consistency for fact verification on their own. In addition, our experiments show that consistency training brings improvements in both settings using only gold evidence. However, the retrieved evidence in real-world setting can be noisy and incomplete. That said, the performance of ProToCo on non-oracle evidence requires further study. To utilize consistency constraints, ProToCo still needs to fine-tune the PLMs. Also, in zero-shot setting, the labels of logical variants are assigned with the predictions of the original claim by the base model, which could be inaccurate and thus affect the consistency training.

\bibliography{anthology,custom}
\bibliographystyle{acl_natbib}

\appendix

\end{document}